\documentclass[11pt]{article}

\usepackage{amsmath}
\usepackage{amssymb}

\usepackage[left=2cm,right=2cm,top=2.5cm,bottom=2.5cm]{geometry}

\usepackage{graphicx,epstopdf}
\graphicspath{{figures/}{figures-pdf/}}
\usepackage{caption}
\usepackage{subcaption}
\usepackage{booktabs} 
\usepackage{float}
\usepackage[export]{adjustbox}

\usepackage{soul, color}
\usepackage{xcolor}

\usepackage{listings}

\usepackage[hidelinks]{hyperref}
\usepackage{url}

\usepackage{stackengine}

\usepackage{verbatim}
\usepackage{stfloats}
\usepackage{makecell}

\title{Deep Learning Based Object Tracking in Walking Droplet and Granular Intruder Experiments}
\author{Erdi Kara\thanks{Corresponding Author, \url{erdikara@spelman.edu }},\thanks{Department of Mathematics, Spelman College}, George Zhang\thanks{Department of Applied Mathematics, University of Washington}, Joseph J. Williams\footnotemark[2], Gonzalo Ferrandez-Quinto\footnotemark[2]\thanks{Department of Physics, University of Washington},\\ Leviticus J. Rhoden\thanks{Department of Mechanical Engineering, University of Washington}, Maximilian Kim\thanks{Department of Civil and Environmental Engineering, University of Washington}, J. Nathan Kutz\footnotemark[2],  Aminur Rahman\footnotemark[2]}
\date{}

\begin{document}

\maketitle

\begin{abstract}
We present a deep-learning based tracking objects of interest in walking droplet and granular intruder experiments. In a typical walking droplet experiment, a liquid droplet, known as \textit{walker}, propels itself laterally on the free surface of a vibrating bath of the same liquid. 
This lateral motion stems from the interaction between the droplet and the wave it generates upon successive bounces off the vibrating liquid surface. A walker can exhibit a highly irregular trajectory over the course of its motion, including rapid acceleration and complex interactions with the other walkers present in the bath. 
In analogy with the hydrodynamic experiments, the granular matter experiments consist of a vibrating bath of very small solid particles and a larger solid called \textit{intruder}. Like the fluid droplets, the intruder interacts with and travels the domain due to the waves of the bath but tends to move much slower and much less smoothly than the droplets. When multiple intruders are introduced, they also exhibit complex interactions with each other.
We leverage the state-of-art object detection model YOLO(You Only Look Once) and the Hungarian Algorithm to accurately extract the trajectory of a walker or intruder in real-time. Our proposed methodology is capable of tracking individual walker(s) or intruder(s) in digital images acquired from a broad spectrum of experimental settings and does not suffer from any identity-switch issues. Thus, the deep learning approach developed in this work could be used to automatize the efficient, fast and accurate extraction of observables of interests in walking droplet, granular intruder experiments and similar particle tracking experiments. Such extraction capabilities are critically enabling for downstream tasks such as building data-driven dynamical models for the coarse-grained dynamics and interactions of the objects of interest.


\end{abstract}

\section{Introduction}
\label{intro}

Since the works of Couder and co-workers \cite{CPFB05,PBC2006,CouderFort06}, walking droplets have revealed a wide array of exotic behavior such as quantum-like phenomena \cite{Bush10,HMFCB13,Bush15a,Bush15b,BushOza20_ROPP}, diffusive behavior \cite{TambascoDiffusion}, orbital dynamics \cite{CPFB05,FEBMC10,OHRB14,OWHRB14,HarrisBush14,Tambasco2016,OSHMB}, and nonlinear dynamics and chaos \cite{Gilet14,BCGMN,RJB17,Rahman18,RahmanBlackmore20,Durey2020c,RahmanBlackmoreReview,Valani2021,Valani2022}.  Droplet tracking has been the crucial first step in characterizing the dynamical properties associated with these observations, and the framework we develop in the present manuscript can substantially accelerate coupled experimental-theoretical studies such as that of Valani and Slim and of Choueiri \textit{et al.} \cite{Valani2018,Choueiri2022}.  The least technical way to track droplets is to manually identify the droplet at each timestep and record its location.  However, this requires a prohibitive amount of human input. The primary bottleneck in automating this process is in droplet identification, after which recording the coordinates of the droplets is trivial.

Similar to vibrating fluid baths, pattern-forming phenomena in vibrated baths of granular matter received serious attention by Melo et al. \cite{metcalf_standing_1997} and Metcalf et al. \cite{melo_transition_1994}, who both elucidated transitions between different standing wave patterns. More complex behaviors can form, and the behavior of vibrated granular materials varies richly with the bed depth $F$, the dimensionless shaking parameter $\Gamma = \frac{a \omega^2}{g} $ (where $a$ is the amplitude of shaking acceleration, $g$ is the acceleration due to gravity, and $\omega$ is the angular frequency of shaking), the particle diameter $d$, and other particle properties such as the coefficients of friction and restitution \cite{eshuis_phase_2007}. 

The experiments with granular matter parallel those with hydrodynamic droplets: a much larger particle, known as the \textit{intruder}, is included in the vibrating media, and for certain ranges of $F$ and $\Gamma$, the intruder will travel the experimental domain, its movement both propelled and mediated by the sea of vibrating particles surrounding it.  The terminology of an \textit{intruder} in the granular system stems from the idea of a much larger particle intruding upon a collection of otherwise identical particles. \cite{kudrolli_size_2004}, \cite{knight_experimental_1996}. Generally, the  much larger particle is carried to the surface by convective forces within the granular matter generated by the vibrations \cite{metzger_all_2011}.

With an effective tracking algorithm,  accurate long-time statistics of droplet and intruders can be extracted.  These statistics are often used to compare the dynamics with that of analogous quantum systems.  Sub-optimal tracking algorithms can potentially lead to incorrect statistical assessments and negatively affect the reproducability of experiments. Moreover, the lack of  reproducibility and erroneous statistics can produce misinformed dynamical models of the underlying physical processes. 

Deep neural networks have become the state-of-the-art method for image recognition applications. Yet there have been few studies leveraging deep learning based object detection and tracking methods for applications related to droplet dynamics in micro-fluids and cellular imaging. Some applications of deep learning such as droplet detection, sorting and classification are discussed in \cite{srikanth2021droplet}. In \cite{durve2021fast,durve2022droptrack,durve2021tracking}, YOLO \cite{yolov5paper} combined with the object tracking algorithm Deep SORT \cite{wojke2017simple} is used to track droplet motion in a class of microfluidic experiments and soft granular flows based on synthetic images generated via Lattice-Boltzman simulations. In \cite{rutkowski2022microfluidic}, Faster R-CNN\cite{ren2015faster} and YOLO are compared for droplet detection in several microfluidic experiments. However, to the best of our knowledge, there has been no deep-learning based investigation regarding extraction of walking droplets and granular intruders from real experiments.

The inherent complexities of tracking walking droplets, characterized by their unique, morphing geometries and highly nonlinear motion, present significant challenges. These challenges can be amplified under less than ideal experimental conditions, such as sub-optimal lighting or low-resolution video capture. While our efforts should always aim towards optimizing experimental setups, we recognize that there may be limitations in real-world scenarios. Consequently, there's a clear need for advanced, robust tracking algorithms capable of performing accurately even under these constraints. This becomes even more essential as precise long-time statistics are imperative to distinguish quantum-like
behavior from probabilistic noise.

There are several key differences between the behaviors and tracking of the walking droplets and the granular intruders. First, the larger size of the intruder compared to the hydrodynamic droplet means that we are less vulnerable to false positives. Second, while multiple droplets may collide and merge into a single droplet, the intruders will never merge, but instead collide and bounce apart. Indeed such collision events pose their own challenge for particle-tracking algorithms due to the large velocity and acceleration of the intruders after the collision. However, intruders never merging is convenient for running an experiment for very long periods; in turn, this is necessary as the slow and erratic motion of the intruders means longer experimental time is required in order for the intruder to meaningfully explore the domain. Finally, to our knowledge, there has never been an application of intruder identification and tracking in any similar granular flow experiments involving surface waves.

Lastly, we would like to clarify the \textit{identity switch} concept, which will be refered as \textit{ID switch} throughout the paper. In the context of object tracking, an ID switch refers to a situation where the tracking algorithm mistakenly assigns different identifiers or labels to the same object over time. Essentially, it means that the algorithm fails to maintain a consistent association between the object being tracked and its corresponding label throughout the tracking process. If ID switches occur during the tracking of particles in our experiments, it means that the algorithm wrongly associates different labels with the same particle or assigns the same label to different particles at different times. This leads to fragmented trajectories where the motion of a single particle is erroneously divided into multiple tracks or merged tracks of different particles.  As a concrete example, in a two-frequency driven experiment, Valani \textit{et al.} \cite{Valani2019} observe far more contact interactions with the bath than their single-frequency counterpart, which complicates the tracking problem and exacerbates ID switching.  The algorithm developed in the present manuscript would allow researchers to confidently track such phenomena.

It is crucial to emphasize that our success criteria for walking droplet and granular intruder tracking is the \textit{complete absence of any ID switches}. We consider zero ID switch as the definitive indicator of accurate and reliable tracking. Because, once an ID switch occurs, no matter how few or infrequent, it introduces errors and distort the true trajectories of the particle of interests. It is important to highlight that attempting to repair or correct these erroneous trajectories through post-processing efforts is practically infeasible. Thus, our ultimate  goal is simply to achieve a flawless tracking performance without any ID switch. Adhering to this strict requirement is vital to ensure the integrity and validity of the obtained tracking data. It guarantees that the extracted trajectories faithfully represent the actual motion of walking droplets, enabling meaningful analysis and interpretation of their complex dynamics

The remainder of the manuscript is as follows: In Sec. \ref{sec:datacollect}, we outline the data acquisition stages regarding walking droplet and granular intruder experiments. 
In Sec. \ref{sec:modeltrain}, we detail the preparation of object detection datasets described in Sec. \ref{sec:datacollect} and discuss the model training and testing. Then, in Sec. \ref{sec:results}, we present our detection and tracking results.  Finally, in Sec. \ref{sec:conclusion} we conclude our study with discussions and future directions.

\section{Data Collection}
\label{sec:datacollect}
\subsection{Walking Droplets} 
\label{sec:drop_exp}

While experimenting with previously available droplet videos and detection methods, we noticed that variations in some experimental variables, namely lighting and droplet resolution, led to inconsistent detection accuracy. Therefore, to determine the optimal experimental setup which yields the best performance, we gathered 10 videos of droplets while varying five experimental parameters (lighting, droplet resolution, corral color, forcing amplitude, number of droplets) unilaterally. The specific parameter values used for each video are listed in the Table-\ref{tab:expsX}. 

\begin{table*}[!ht]
  \centering
  \caption{Experimental parameters for walking droplet videos}
  \label{tab:expsX}
  \begin{tabular}{lllccc}
    \toprule
    Experiment &Lighting & Resolution & Color & Amplitude (g) & \# of Droplets \\
    \midrule
    Control & Medium (6) & High & Black & 1.47 & 1 \\
    Lights Off & Off (0) & High & Black & 1.47 & 1 \\
    Lights Low & Low (1) & High & Black & 1.47 & 1 \\
    Lights High & High (11) & High & Black & 1.47 & 1 \\
    Two Droplets & Medium (6) & High & Black & 1.47 & 2 \\
    Three Droplets & Medium (6) & High & Black & 1.47 & 3 \\
    Res Low & Medium (6) & Low & Black & 1.47 & 1 \\
    Res Mid & Medium (6) & Medium & Black & 1.47 & 1 \\
    Faraday & Medium (6) & High & Black & 1.53 & 1 \\
    Corral White & Medium (6) & High & White & 1.47 & 1 \\
    \bottomrule
  \end{tabular}
\end{table*}

All videos were captured with an iPad Pro mounted on a tripod at 720p (which are all 30fps except for Corral White which was 60fps). To reduce algorithmic workload and minimize distractions, the videos were cropped to only contain the corral region.

Experimental variables not listed were kept controlled to the best of our abilities. We first 3D printed identically-shaped black and white circular corrals using Polylactic Acid (PLA). Then, we filled the corrals to the same height for each experiment with a $20cSt$ silicone oil. The forcing frequency was also kept at 60Hz for all experiments.

Different levels of lighting were achieved with a tabletop LED photography light as the only light source within the room. The LED light has 11 different illumination levels, of which we captured footage at level 0 (Lights Off), level 1 (Lights Low), level 6 (Lights Mid), and level 11 (Lights High). Varying droplet resolution is achieved by placing the iPad at different vertical distances from the corral. The closer the iPad is to the corral, the higher the droplet resolution.

Both the silicone oil fluid bath and the droplets on top appear and behave drastically different depending on whether the forcing is below or above the Faraday threshold. Thus, we kept the frequency of the forcing constant at 60Hz and varied the amplitude to stay just below the Faraday threshold (1.47g) and to exceed the threshold (1.53g).

\subsection{Granular Materials}

The granular matter experiments share some similarities to the walking droplets experiments, but differ in some key ways. First, the bath of smaller particles consists of chrome steel bearing balls of $d=1.6$ mm and a bed depth of $ F = 7.5 $ mm. The bath is enclosed in a cylindrical container with $ D = 214 $ mm. We use intruder particles of two different ceramics, silicon 
nitride ($ \mbox{Si}_{\mbox{3}} \mbox{N}_{\mbox{4}} $) and zirconium oxide ($ \mbox{Zr} \mbox{O}_{\mbox{2}} $), both less dense than chrome steel. The intruders are sized $ d = 9.5 $ mm.
    
Key differences in the experiments are found in the hyper-parameters. Neither lighting nor resolution were hyper-parameters for the granular matter experiments, but intruder color was. The $ \mbox{Si}_{\mbox{3}} \mbox{N}_{\mbox{4}} $ ceramic is white and the $ \mbox{Zr} \mbox{O}_{\mbox{2}} $ ceramic intruders are black. Identity switching is more likely when intruders of the same color are interacting with each other. The color of the corral was not varied. Refer to Table-\ref{tab:Granular_Experiments} for the experimental parameters of our granular systems.

\begin{table*}
\centering
\caption{Experimental parameters for granular intruders videos}
\label{tab:Granular_Experiments}
\begin{tabular}{llllll}
    \toprule
    Experiment & Intruder Material & Intruder Color & Amplitude (g) & Frequency $f$ & \# of Intruders \\
    \midrule
    3white              & 3x $ \mbox{Si}_{\mbox{3}} \mbox{N}_{\mbox{4}} $  & 3x White             & 2.26 & 20 & 3 \\
    2white2black-long   & 2x $ \mbox{Si}_{\mbox{3}} \mbox{N}_{\mbox{4}} $, 2x $ \mbox{Zr} \mbox{O}_{\mbox{2}} $ & 2x White, 2x Black   & 2.15 & 20 & 4 \\
    2white2black-short  & 2x $ \mbox{Si}_{\mbox{3}} \mbox{N}_{\mbox{4}} $, 2x $ \mbox{Zr} \mbox{O}_{\mbox{2}} $ & 2x White, 2x Black   & 3.10 & 50 & 4 \\
    \bottomrule
\end{tabular}
\end{table*}

As with the walking droplets, the granular experiments were also filmed with an iPad Pro mounted on a tripod at 720p and 30fps and the videos were cropped. The frequency and amplitude of the vibrations were varied across the experiments to ensure sufficient wave patterns and intruder activity.

\section{Deep Learning Based Object Detection}
\label{sec:modeltrain}

As a fundamental computer vision problem, object detection is the task of locating objects and categorizing them into predefined classes in images or video frames. Autonomous driving, video surveillance systems, and real-time scene understanding are just a few applications of object detection.

Conventionally, object detection has been achieved without neural networks and deep learning. Image processing techniques such as the Hough transform and edge detection extract information from images by applying pixel-level operations or filters \cite{10.1145/361237.361242,YUEN199071,ATHERTON1999795,thapar2012study,mathworks-2011,4767851,MEYER1994113}. Background subtraction excels in extracting moving foreground objects in front of static backgrounds \cite{https://doi.org/10.48550/arxiv.1302.1539,8652040,Stauffer1999AdaptiveBM,KaewTrakulPong2002AnIA,1333992,ZIVKOVIC2006773,grosek2014dynamic,erichson2019compressed}. Machine learning approaches such as the Viola-Jones algorithm train to detect objects given labelled data \cite{990517}. Successful implementations of image processing \cite{plantwilt,fruitsontree,9665023,KULJU201848,8621816,ADM}, background subtraction \cite{henstracking,10.1371/journal.pone.0239504}, and more recently machine learning techniques \cite{4541129} have been readily reported in agricultural and biological contexts and, to our interest, on various forms of droplets. Readers can consult \cite{szeliski2022computer,forsyth:hal-01063327,5961} for a more comprehensive review of commonly applied computer vision methods. While these techniques are not the main focus of our paper, we acknowledge their ability and viability. Thus, see Appendix A for the detection results of these methods on walking droplets.  In addition, we note that the background subtraction problem typically assumes a static background.  However, the droplet and intruder system both have objects of interest whose background is also dynamics.  For instance, the granular material is dynamic as the intruder evolves over its trajectory.  This makes the detection problem more challenging.

As the complexity and amount of experimental data increases, conventional feature extraction processes have difficulty providing effective detection and tracking results. In the last 10 years, deep neural networks achieved breakthrough results particularly in image recognition domain. In terms of performance, many of these conventional methods were surpassed by neural networks in almost all image recognition applications including object detection. Since the full review of deep learning based object detection literature is beyond the scope of this paper, readers may consult to \cite{zaidi2022survey,liu2020deep} for an extensive review of current state of this field.

In this work, we use the YOLO architecture proposed by Redmon et.al in \cite{redmon2016you}. Since then, it has been extensively adopted in computer vision research due to its fast and accurate object detection capabilities in real time. In the subsequent years, the model was gradually improved resulting in several versions \cite{yolov2,yolov3,yolov4} and achieved state-of-art results on the most commonly used object detection datasets such as Pascal VOC and COCO \cite{coco,pascal}.  

In this paper, we use mean average precision (mAP), a commonly used metric for evaluating object detection models, as the evaluation metric of our models. mAP is defined via \textit{intersection-over-union} (IOU) that measures the degree of overlap (0 being no overlap, 1 being complete overlap) between the bounding box prediction and the ground truth bounding box. mAP . mAP measures the average precision (AP) across multiple IOU thresholds (typically from 0.5 to 0.95) for a set of object classes. 
We will report mAP@0.5 which refers to the mAP calculated using a single IOU threshold of 0.5. We will also report mAP@[.5:.95] which refers to the mAP calculated using multiple IOU thresholds ranging from 0.5 to 0.95, with a step size of 0.05. For a more detailed explanation of these metrics, readers can refer to supplementary materials.

\subsection{Data Preparation}
Table--\ref{tab:train} details the properties of each video regarding the walking droplet and granular intruder experiments introduced in Sec-\ref{sec:drop_exp} as well as the number of training, validation and testing images utilized to train and test the YOLO architecture described above.

\begin{table*}[!ht]
    \centering
    \caption{Video characteristics of each walking droplet and granular flow experiments with number of frames used for model training and testing}
    \label{tab:train}
    \begin{tabular}{llllll}
    \toprule
        Experiment & Duration(mins) & FrameCount & TrainImg & ValidImg & TestImg \\ 
        \midrule
        Control & 4.31 & 7494 & 120 & 34 & 17 \\ 
        Lights off & 4.53 & 7884 & 126 & 36 & 18 \\ 
        Lights low & 4.58 & 7962 & 127 & 36 & 18 \\ 
        Lights high & 4.27 & 7426 & 124 & 36 & 17 \\ 
        Two droplets & 4.62 & 8032 & 129 & 36 & 18 \\ 
        Three droplets & 5.19 & 9034 & 127 & 36 & 18 \\ 
        Res mid & 5.63 & 9798 & 123 & 35 & 17 \\ 
        Res low & 4.49 & 7805 & 125 & 36 & 17 \\ 
        Faraday & 4.67 & 8118 & 130 & 37 & 18 \\ 
        Corral White & 2.08 & 7692 & 126 & 36 & 17 \\
        3white & 7.04 & 7884 & 31 & 8 & 4 \\
        2white2black-long & 20.22 & 36259 & 31 & 8 & 4 \\
        2white2black-short & 10.31 & 17893 & - & - & - \\ 
        \bottomrule
    \end{tabular}
\end{table*}

Our general approach for walking droplet experiments is to employ around 180 frames that are sampled with uniform time intervals from the corresponding video source. Out of these frames, we then reserve 70\% for training, 20\% for validation, and 10\% for testing. Notice that we consider only around 2\% of the frames relative to the total number of frames present in the video sources. Since data preparation takes up the largest chunk of time in model training, it is essential to keep the training data as small as possible without compromising the model accuracy. In particular, the droplets in our dataset are approximately squares 6-10 pixels wide which renders manual bounding box annotation particularly challenging. But using more training images does generally result in better performing models. Readers can refer to Section 2 of the supplementary material to learn about the specific details of how we selected the number of images used to train our model.

Notice that we employ considerably less images compared to droplet experiments. As it will be detailed below, large and distinct shapes of intruders relative to their surrounding medium significantly helps the YOLO model to learn from a small number of training samples. Due to the same reason, we do not create a separate annotations for the 2white2black-short experiment since the intruders in both experiments are qualitatively similar.  

As noted above,  object detection datasets are usually composed of images and the coordinates of the bounding boxes associated with each object of interest in the corresponding image or frame. To create our dataset, we used the online annotation tool \textit{LabelImg} \cite{tzutalin2015labelimg}.

\vspace{-10pt}
\subsection{Model Training}
In this work, we adopted the Pytorch  implementation of the YOLO architeture  maintained by Ultralytics in \cite{yolov8}. In particular, we selected the \textit{YOLOv8} architecture which is the most recent implementation of YOLO series. Among the existing YOLOv8 networks, we will work with the most lightweight one known as \textit{yolov8n}. We can train a separate model for each of the experiments above or a single model by combining the individual datasets. For simplicity, we use the single model option. However, we experimented with the former as well and observed similar results in both cases. 

While the combined walking droplet dataset is composed of 1257 training, 358 validation and 175 testing images with their annotations, the granular intruder dataset has 62 training, 16 validation and 8 testing images. We trained the models with pretrained COCO weights and used the default set of hyperparameter configuration provided by the same repository \cite{yolov8}. Training and testing results regarding these experiments are summarized in Table-\ref{tab:train_results}.

\begin{table*}[!ht]
    \centering
    \caption{YOLOv8 training and testing metrics for both experiments}
    \label{tab:train_results}
    \begin{tabular}{*{7}{l}}
    \hline
        Experiment & Time & Epochs & \multicolumn{2}{l}{Training} & \multicolumn{2}{l}{Testing} \\ \cline{4-7}
        ~ & ~ & ~ &                  mAP@.5 & mAP@[.5:.95] & mAP@.5 & mAP@[.5:.95] \\ \hline
        Walking Droplet & 37 & 253 & 0.995 &  0.561 &       0.98 &   0.489 \\
        Granular Flow & 4.6 & 186 & 0.995 &   0.722 &       0.995 &  0.678 \\ \hline
    \end{tabular}
\end{table*}

All training and testing stages were performed on a GE66 Raider 10SFS with a 2.60GHz Intel Core i7-10750H CPU and an NVIDIA GeForce RTX 2070 Super GPU. 

\section{Results}
\label{sec:results}

\subsection{Walker and Intruder Detection}
\label{sec:detection}
Once we have a trained model accurately detecting droplets, we can process the walking droplet and granular intruder videos on a frame-by-frame basis and detect the walker or intruder locations in each frame. We perform this operation in real-time. For each frame in the corresponding video source, we accept a detection is valid only if the detected number of droplets or intruders is equal to the number of droplet(s) or intruder(s) in the related experiment. Moreover, we strictly impose that the confidence score associated with each droplet or intruder detection must be above a certain threshold value. In this work, we set this threshold to 0.45. For example, if we consider three-droplet experiment, we accept a detection valid only if the model detects exactly three droplets with each confidence score above 0.45. Note that since the size of the droplet is small compared to the actual frame and the droplet itself is composed of the same fluid as the medium, one may encounter false-positives either on the medium or even on the experiment tray. Although this effect appears to be less pronounced in granular experiments, we would like to maintain relatively high confidence for each intruder as well.  As detailed in the next section, one of the primary goals of this paper is to extract the individual droplet or intruder trajectories. Such a criteria helps us avoid false-positive identifications which essentially renders the extracted trajectory unreliable. Using this approach, we count how many frames fulfill this criteria and define \textit{frame detection rate} (FDR) as the ratio of frames passing this criteria to the total number of frames. The results in Table-\ref{tab:ftr1}  demonstrate a near perfect performance of the YOLOv8 model. Among 13 experiments we conducted, there was no instance where a
false-positive identification(i.e a background object identified as droplet or intruder)managed to circumvent this our criterion. 

\begin{table*}[!ht]
     \centering
    \caption{Frame detection rates (FDR) for walking droplet and granular intruder experiments.}
    \label{tab:ftr1}
    \begin{tabular}{lccccc}
    \toprule
        Experiment & Detections & \makecell{Total\\Frames} & \makecell{Missed\\Frames} & FDR & \makecell{Inference\\Time (sec)} \\
        \midrule
        Control & 7494 & 7494 & 0 & 1 & 94 \\ 
        Lights Off & 7883 & 7884 & 1 & 0.99987 & 101 \\ 
        Lights Low & 7961 & 7962 & 1 & 0.99987 & 101 \\ 
        Lights High & 7424 & 7425 & 1 & 0.99987 & 94 \\ 
        Two Droplet & 7978 & 8031 & 53 & 0.9934 & 121 \\ 
        Three Droplet & 8909 & 9033 & 124 & 0.98627 & 160 \\ 
        Res Mid & 9795 & 9797 & 2 & 0.9998 & 125 \\ 
        Res Low & 7801 & 7805 & 4 & 0.99949 & 93 \\ 
        Faraday & 8069 & 8118 & 49 & 0.99396 & 104 \\ 
        Corral White & 7581 & 7692 & 111 & 0.98557 & 137 \\ 
         3white & 12569 & 12720 & 151 & 0.98813 & 274 \\ 
        2white2black-10m & 18633 & 18917 & 284 & 0.98499 & 563 \\ 
        2white2black-20m & 36506 & 36663 & 157 & 0.99572 & 1831 \\ 
        \bottomrule
    \end{tabular}
\end{table*}


Our proposed methodology achieves near perfect detection rates across a broad spectrum of experimental settings. While it is possible to construct extreme experimental conditions to test the limits of any model, our primary focus remains on producing accurate trajectories of walking droplets and granular materials under realistic and optimal conditions. This includes maintaining even lighting and utilizing a stable, high-resolution camera setup. Nevertheless, we acknowledge that real-world experiments may be subject to external noise and unforeseen challenges. To this end, we demonstrate that our methodology is robust enough to handle the adverse circumstances encountered, such as lighting conditions, variability in resolution and frame rate, and the nonlinear motion of the objects.

The near perfect detection presented in Table-\ref{tab:ftr1} will constitute the most important component of the trajectory extraction task which is the overarching goal of this paper. To emphasize the importance of accurate detection, let us consider the Three Droplets experiment where the model is capable of detecting 8909 out of 9034 frames. The corresponding experiment video is captured approximately 30 frame-per-second (fps) with a length of 5.19 minutes. This means the model is missing only 4.1 seconds out of 5.19 minutes. Similarly, the model is missing only 5.2 seconds out of 22.22 minutes in the 2white2black-long granular experiment. We carefully note that those undetected frames are scattered across different times in the video (see Fig-3 in supplementary material). Thus, its effect on trajectory extraction is essentially negligible. The reason behind a frame not meeting the criteria is usually because the detected number of particles doesn't match the actual number of particles.  For instance, in Three Droplet experiment, out of the 124 frames missed (a small fraction of the total number of frames), the model detected more than three particles in 100 of these frames and detected only two in the remaining 24. We should note that among 13 experiments we conducted, there was not a single instance where a false-positive identification managed to circumvent this stringent criterion. Maintaining a near perfect frame detection rate is particularly important for experiments including multiple walkers or intruders. This point will be detailed in Sec-\ref{sec:tracking}.

\begin{figure}[!ht]
\centering
\begin{subfigure}{0.45\textwidth}
  \stackinset{l}{1mm}{t}{1mm}{\LARGE \color{white} \textbf{a}}{\includegraphics[width=\linewidth]{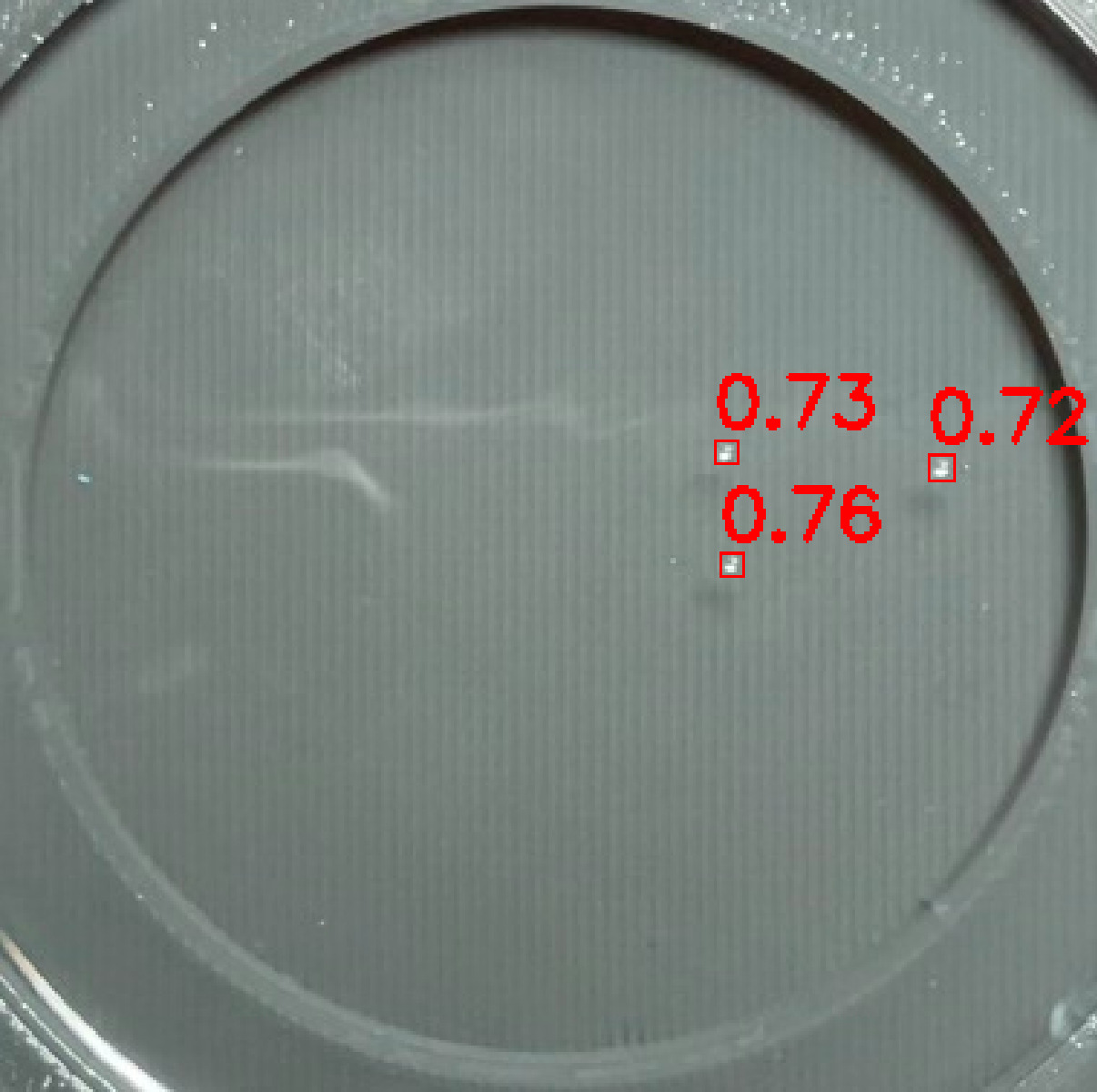}}
\end{subfigure}\hspace{1mm}
\begin{subfigure}{0.45\textwidth}
   \stackinset{l}{1mm}{t}{1mm}{\LARGE \color{white} \textbf{b}}{\includegraphics[width=\linewidth]{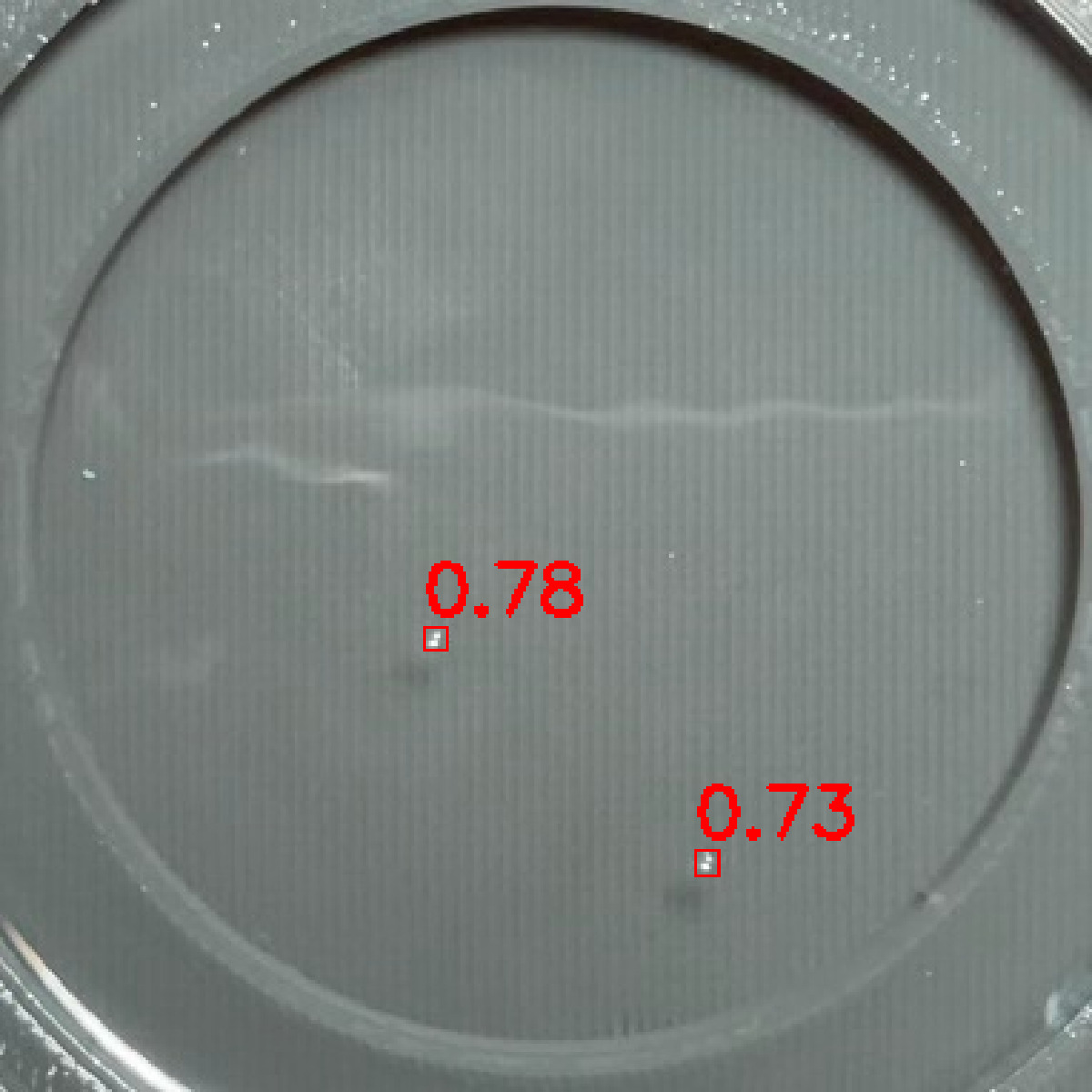}}
\end{subfigure} 
\caption{Walking droplets detection samples from our implementation of YOLOv8.  The red solid square indicates YOLOv8 finding the walker, and the decimal values next to the square indicate the confidence score in the range of $0$ to $1$. \textbf{(a)}, \textbf{(b)} and \textbf{(c)} videos selected from the test dataset of the "Three Droplets" and "Two Droplets", respectively.}
\label{fig:detection}
\end{figure}

Some detection samples with bounding boxes and their associated confidence scores are displayed in Fig-\ref{fig:detection}. Samples in the top rows are from the test dataset of the Three Droplets, Two Droplets and Lights Off experiments, respectively. In each instance, the droplets are accurately detected with high confidence.

Notice the usage of black corrals in all of the experiments except one. We carried out one experiment with a white corral holding a walker in an attempt to test the performance of YOLOv8 in a more demanding setting. As opposed to the previous experiments which are all 30 fps, we also captured this experiment at 60 fps. While this task presents an elevated level of difficulty due to reduced contrast between the walker and its background - a scenario that might pose a heightened challenge for human observers - our method remains capable and efficient in these circumstances by capturing 99\% of the frames.

\begin{figure}[htbp]
    \centering 
\begin{subfigure}{0.45\textwidth}
  \stackinset{l}{1mm}{t}{1mm}{\Large \textbf{a}}{\includegraphics[width=\linewidth]{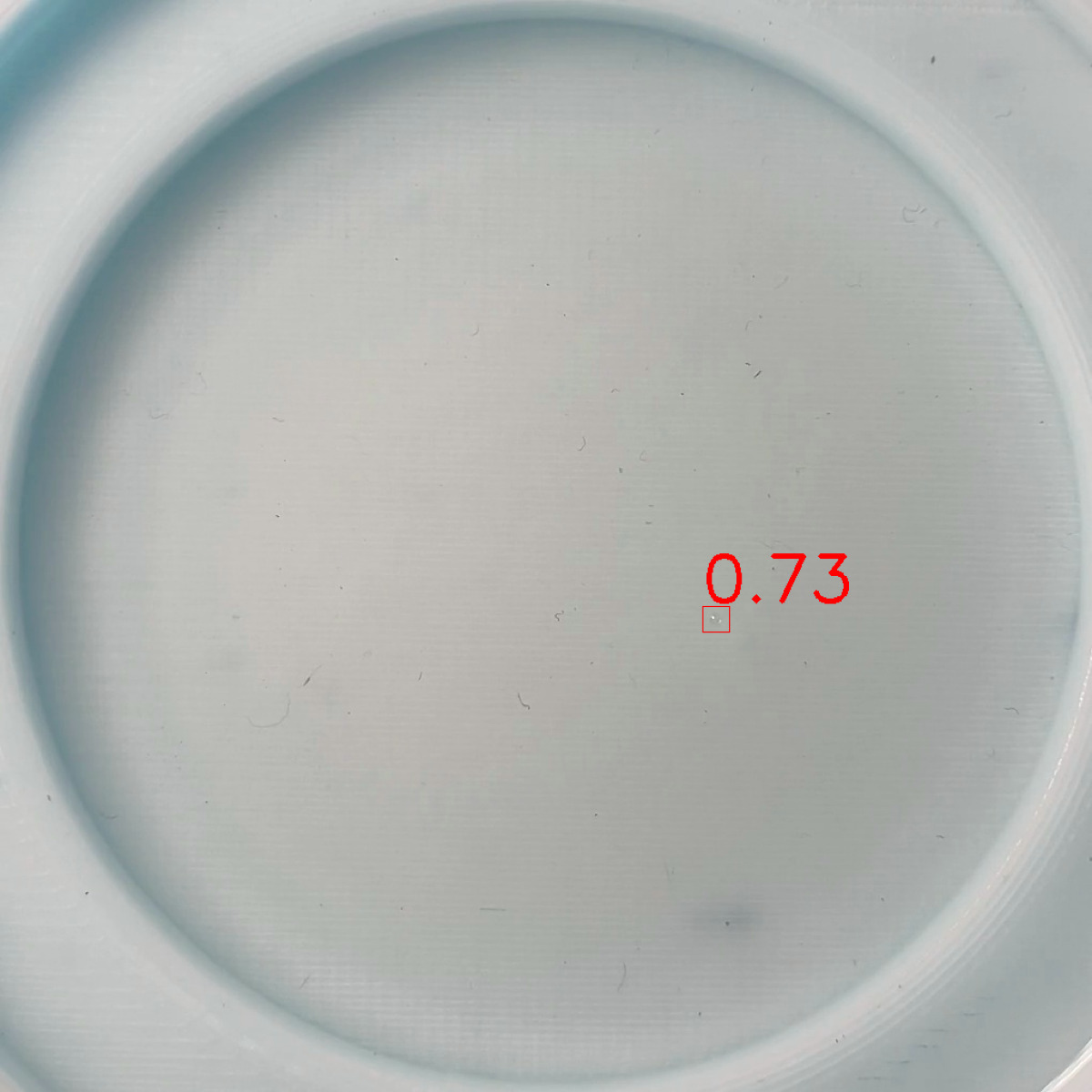}}
\end{subfigure}\hspace{1mm} 
\begin{subfigure}{0.45\textwidth}
  \stackinset{l}{1mm}{t}{1mm}{\Large \textbf{b}}{\includegraphics[width=\linewidth]{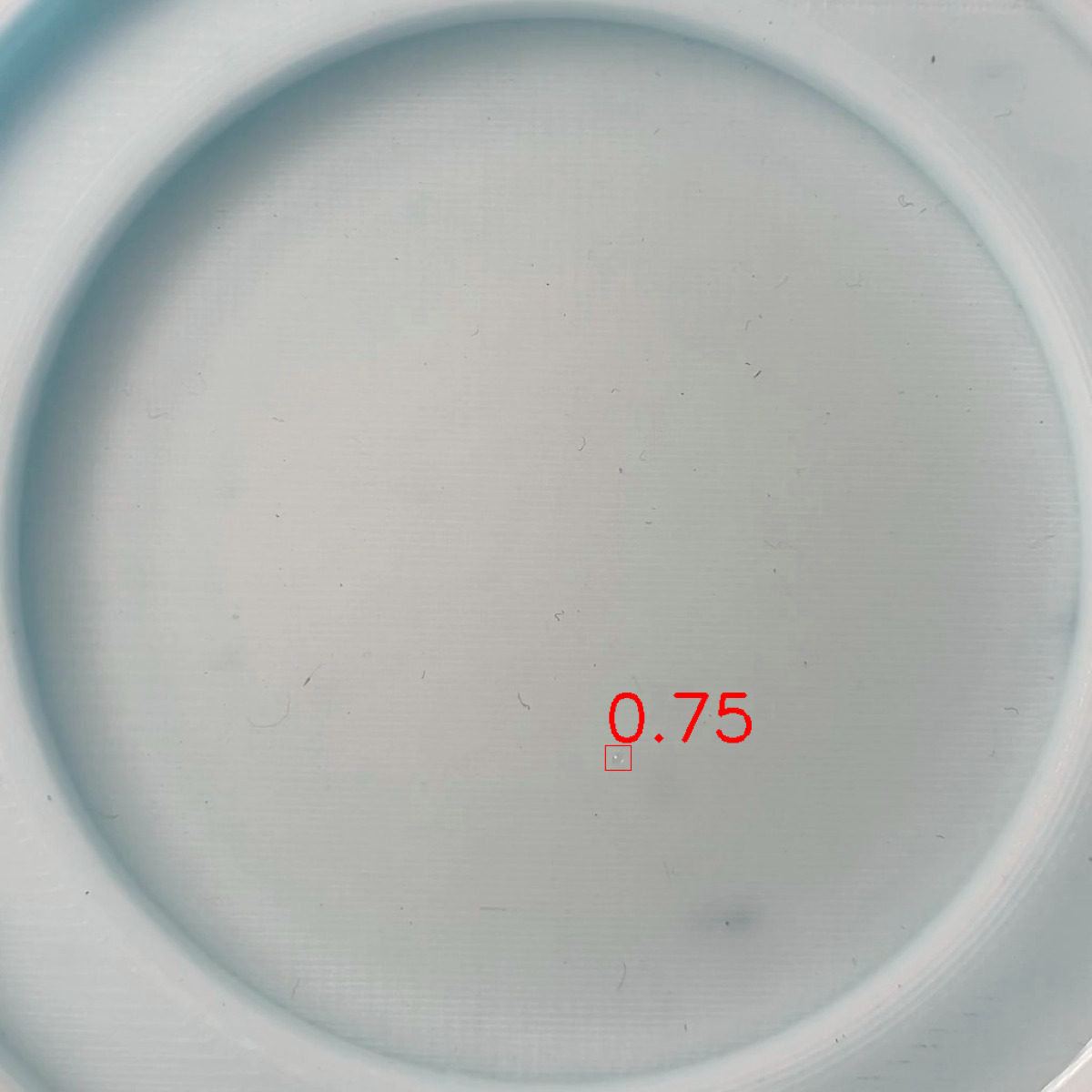}}
\end{subfigure} 
\caption{The walker captured by YOLOv8 in the ``White Corral'' video in two separate frames.  The red solid square indicates YOLOv8 finding the walker, and the decimal values next to the square indicate the confidence score in the range of $0$ to $1$.}
\label{fig:white}
\end{figure}

\begin{figure}[!ht]
    \centering
\begin{subfigure}{0.45\textwidth}
    \stackinset{l}{3mm}{t}{3mm}{\LARGE \color{white} \textbf{a}}{\includegraphics[width=\linewidth]{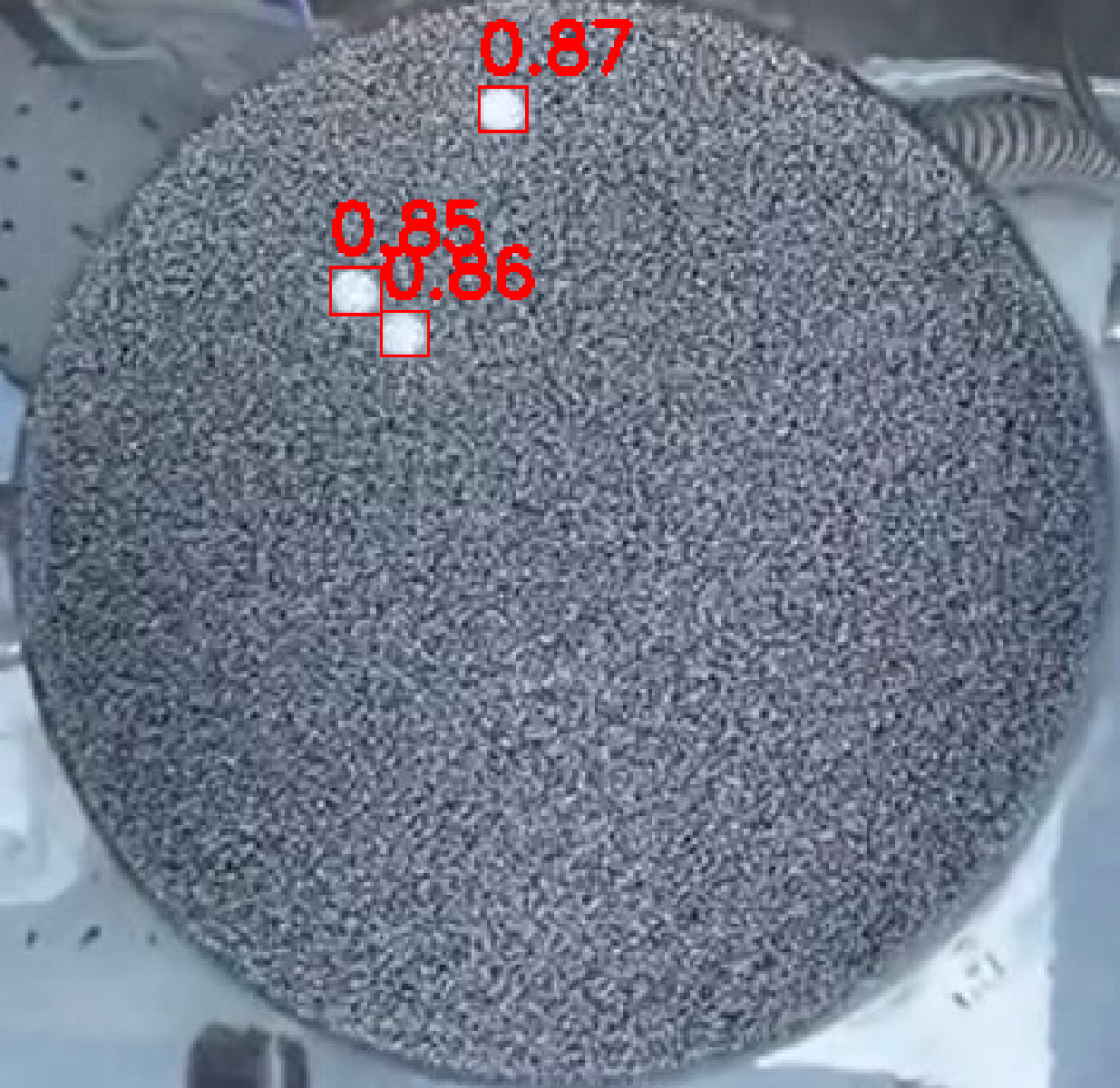}}
\end{subfigure}\hspace{1mm}
\begin{subfigure}{0.45\textwidth}
    \stackinset{l}{3mm}{b}{3mm}{\LARGE \color{white} \textbf{c}}{\includegraphics[width=\linewidth]{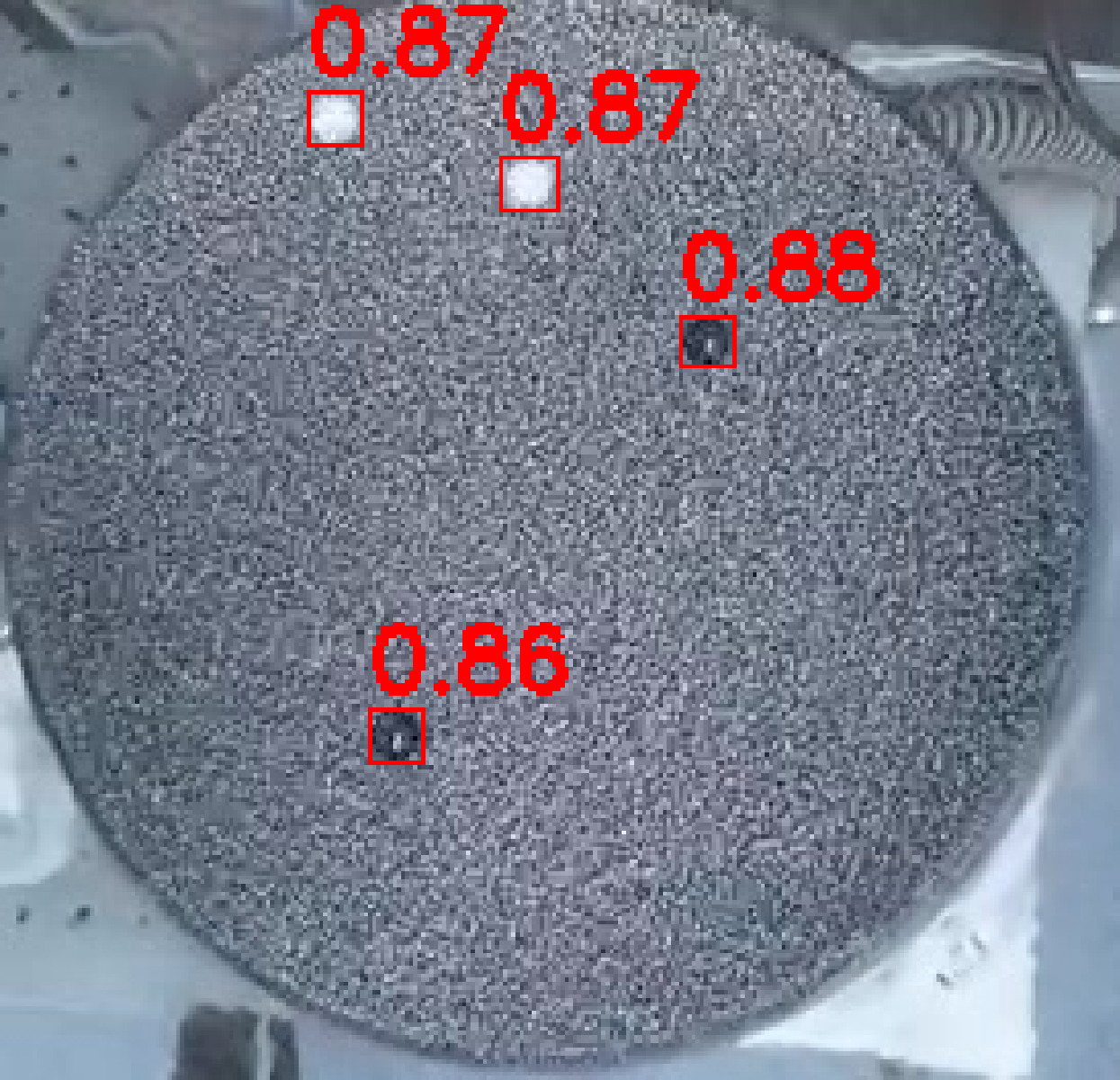}}
\end{subfigure}\hspace{1mm}
\caption{Granular material detection samples by our YOLOv8 model from experiments with three white intruders (\textbf{a}) and two white and two black intruders (\textbf{b}).  The red solid square indicates YOLOv8 finding the intruder, and the decimal values next to the square indicate the confidence score in the range of $0$ to $1$.}
\label{fig:detectint}
\end{figure}

Similarly, the model can detect the intruders in the granular material experiments with high confidence. We display some samples in Fig-\ref{fig:detectint}.

\subsection{Trajectory Extraction by Object Tracking }
\label{sec:tracking}

Object tracking refers to assigning unique identities to objects in an image and tracking them in the subsequent frames. In computer vision, this task is defined as multiple object tracking (MOT) or single object tracking (SOT) depending on the task. The essential step in object tracking is first detecting the objects. Thus, all the challenges present in object detection problems inherently exist for object tracking as well. Occlusion, nonlinear motion, appearance or disappearance of objects between the frames and similar visual features of the objects are just a few difficulties that make object tracking a challenging task. Failure of a tracker can manifest itself in different forms. Trackers may lose an already tracked object, confuse the initial object identities with other class identities in the subsequent frames resulting in \textit{identity switches (ID switch)} or fail to track the objects all together. If the ultimate goal of the study is to accurately identify the trajectory of an object, the tracker must be free of these defects to a great extent. Since the development of a new MOT framework is not the focus of this work, the reader is encouraged to consult excellent review articles which give a comprehensive review of MOT methods and approaches~\cite{pal2021deep,milan2013challenges,ciaparrone2020deep}. Among the existing methods, the most common approach is tracking-by-detection which works in two stages. In the first stage, the object of interests are located in the scene. In the second stage, these object are associated with already tracked ones by a tracking model. Most of the MOT algorithms are equipped with some or all of the following features; (1) detection, (2) feature extraction (e.g with the help of neural networks) or motion prediction (e.g Kalman Filter), (3) affinity and (4) association \cite{ciaparrone2020deep}.

We demonstrated in the previous section the high performance of our YOLOv8 model for various forms of experiments in consideration. As it is the case in many machine learning tasks, it is essential to integrate domain specific knowledge into the problems we tackle. In the context of our experimental settings, the number of walker or intruders essentially dictate the number of objects detected per frame since disappearance of an existing walker or intruders or appearance of new ones are not allowed. Therefore, false-positive identifications can be handled to a great extend by only considering frames where the detection count matches the actual walker or intruder count and imposing a high confidence score on detections. However, as detailed below, high and accurate detection rates may not guarantee accurate tracking even if one employs a state-of-the-art object tracking framework.

Given the accurate detection performance of YOLOv8, the task is to build a tracker which is capable of tracking each individual walker or intruder in our experiments. Several issues remain post-detection that render droplet or intruder tracking a very challenging task. Droplets in our experiments are only 7-10 pixels across, lack rich visual context and exhibit highly nonlinear motion, especially when they collide. A similar challenge is also present in granular intruder experiments as the shape and visual properties of intruders are quite similar to each other. The small size and almost identical appearance of the walkers or intruders combined with chaotic motion are fundamental obstacles that may hinder the performance of the tools employed in 
feature extraction and motion prediction stage of MOT algorithms. Even with deep neural networks, feature extraction cannot assist in tracking given nearly indistinguishable droplets. Motion detection fails and can even lead to identity switches when the motion of walkers or intruders follow no perceivable pattern.

At this point, we can turn our attention to one of the state-of-the-art tracking libraries and implement some of them for our problem. However, we argue that one can employ a solely motion based tracking method to solve this problem. In this sense, we employ the Hungarian Algorithm which solves the linear assignment problem (sometimes called 2-assignment problem) in polynomial time \cite{kuhn1955hungarian}. Let $\big\{c_{i}^{(f)}\big\}_{i=1}^{n}$ and $\big\{\hat{c}_{i}^{(f+1)}\big\}_{i=1}^{n}$ be the center of the bounding boxes representing the location of $n$ droplets/intruders detected in two adjacent frames $f$ and $f+1$. In literature, $\big\{c_{i}^{(j)}\big\}_{i=1}^{n}$ and  $\big\{\hat{c}_{i}^{(j+1)}\big\}_{i=1}^{n}$ are sometimes called \textit{tracks} and \textit{detections}, respectively. We define the following cost matrix based on Euclidean distance 
\begin{equation}
\label{eq:norm}
   C(i,j) = \| c_{i} - \hat{c}_{j} \| \mbox{ for } i,j=1,2,..n 
\end{equation}
Hungarian algorithm aims to identify $n$ unique indices from $C$ in a way that there is one index in each row and in each column such that the total cost associated with these $n$ indices is minimized. Once these unique indices are identified, they are added to the tracks and the process is repeated. In the actual implementation of the Hungarian Algorithm, we utilized the Scipy library \cite{2020SciPy-NMeth}. The Hungarian Algorithm can be considered as one of the most simplistic approach for MOT problems. Due to its simplicity, it is regarded as a typical base model in object tracking benchmarks.  

In spite of its simplicity, the Hungarian Algorithm is well suited to our problem setting. As noted above, walkers and intruders in our experiments have almost identical visual appearances. Therefore, the feature extraction stage used by many MOT methods may fail to differentiate those similar objects from one scene to the next. Moreover, motion predictor functionalities in MOT frameworks do not perform well if the object of interests exhibit highly nonlinear motion between consecutive frames. Therefore, we argue that the Hungarian Algorithm, which is not corrupted by potentially negative effects of feature extraction and motion prediction, is an ideal option to track the individual walkers and intruders. 

\begin{figure}[!ht]
    \centering
\begin{subfigure}{0.22\textwidth}
    \stackinset{l}{3mm}{t}{3mm}{\LARGE \color{white} \textbf{a}}{\includegraphics[width=\linewidth]{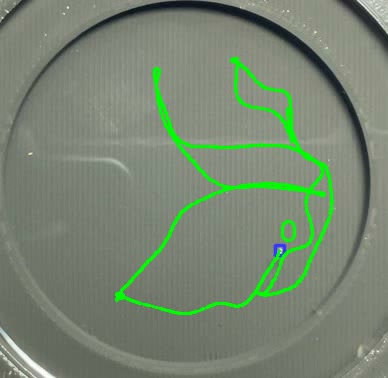}}
\end{subfigure}\hspace{1mm}
\begin{subfigure}{0.22\textwidth}
    \stackinset{l}{3mm}{t}{3mm}{\LARGE \color{white} \textbf{b}}{\includegraphics[width=\linewidth]{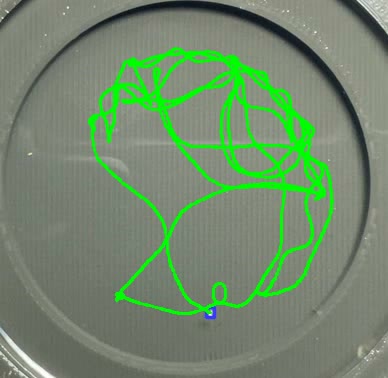}}
\end{subfigure} 
\medskip 
\begin{subfigure}{0.22\textwidth}
    \stackinset{l}{3mm}{b}{3mm}{\LARGE \color{white} \textbf{c}}{\includegraphics[width=\linewidth]{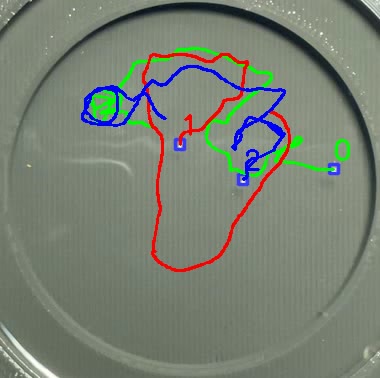}}
\end{subfigure}\hspace{1mm}
\begin{subfigure}{0.22\textwidth}
    \stackinset{l}{3mm}{b}{3mm}{\LARGE \color{white} \textbf{d}}{\includegraphics[width=\linewidth]{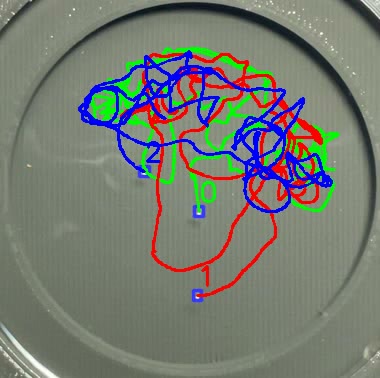}}
\end{subfigure}
\caption{Real time tracking of walking droplets from the Control (\textbf{a, b}) and Three Droplets (\textbf{c,d}) experiments at two different time snapshots. The different colors represent the different individual walkers.}
\label{fig:realtime1}
\end{figure}

We first present the tracking results regarding the single and multiple droplet cases and extract important information to further analyze the characteristics of droplet motion. Our primary goal is to keep track of the coordinates of the bounding box centers for each individual droplet. In particular, we would like to perform this task in real-time. We present two snapshots from the Control experiments in the top row of 
Fig-\ref{fig:realtime1}. Since there is only a single droplet in this experiment, an initial ID = 0 is assigned to the droplet and the same ID is tracked in the subsequent scenes. Since our approach is purely motion based, there is no ID switch in any of the single droplet experiments. We also note again that we are capable of tracking 100\% of the 7494 frames in the Control experiment by using only around 120 training frames for each experiment. For all the other experiments, the reader is highly encouraged to watch the real-time tracking videos provided in the supplementary material.

Note that single droplet tracking is not challenging as long as the detection algorithm accurately identifies the location of the droplet. Once a single location is identified, association with the previous frame is straightforward since there is no another droplet present in the frame. However, multiple droplet experiments pose several characteristics which renders tracking very challenging. One particular challenge is the nonlinear nature of the droplet motion that may include rapid accelerations from one frame to the next. In particular, the moment when two or more walkers approach each other, they generally experience a push stemming from the superposition of pilot waves. This in return causes walkers to rapidly accelerate between two consecutive frames. To avoid ID switches, we rely on the ability of YOLOv8 to resolve the motion, i.e. it keeps detecting the droplets with high confidence. Similar to single droplet experiments, we have not observed any ID switches in multiple droplet experiments. Fig-\ref{fig:realtime1} demonstrates two snapshots captured from Three Droplets experiment. For ease of visualization, we use a different color for each track representing the trajectory of individual droplets. As mentioned above, these results are essentially observed in real time with no external post-processing. In other words, we save and display the trajectories simultaneously until the end of the corresponding experiment. As expected, the trajectories appear to be chaotic rather than following a regular path.

\begin{figure}[htbp]
    \centering 
\begin{subfigure}{0.45\textwidth}
  \stackinset{l}{2mm}{t}{2mm}{\LARGE \color{white} \textbf{a}}{\includegraphics[width=\linewidth]{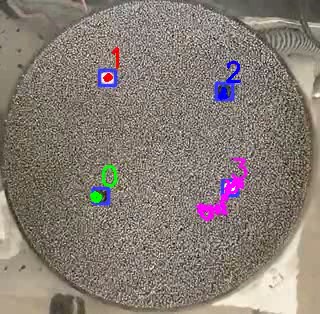}}
\end{subfigure}\hspace{1mm} 
\begin{subfigure}{0.45\textwidth}
  \stackinset{l}{2mm}{t}{2mm}{\LARGE \color{white} \textbf{b}}{\includegraphics[width=\linewidth]{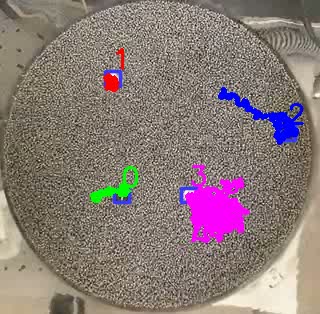}}
\end{subfigure}\hspace{1mm} 
\caption{Real time tracking snapshots from the two white - two black \textbf{(a, b)} granular intruder experiments. The different colors represent the different individual intruders.}
\label{fig:realtime3}
\end{figure}

Similarly, we can extract the trajectories of individual intruders in granular flow experiments. The results in Fig-\ref{fig:realtime3} indicate that the intruders exhibit significantly less displacement than walking droplets. 

\begin{figure}[!ht]
    \centering
\begin{subfigure}{0.39\textwidth}
    \stackinset{l}{4mm}{t}{-3mm}{\huge \textbf{a}}{\includegraphics[width=\linewidth]{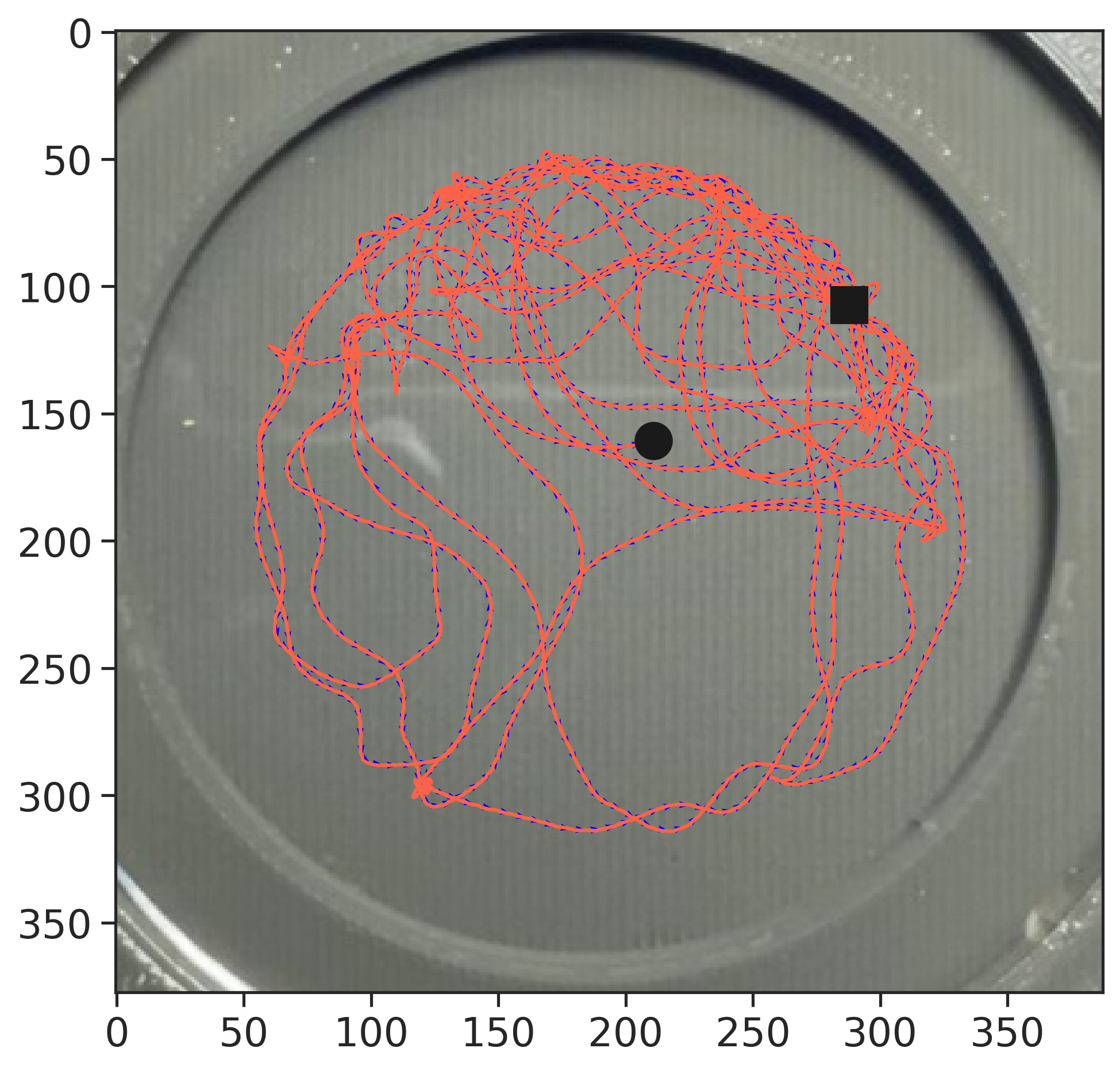}}
\end{subfigure}\hspace{1mm}
\begin{subfigure}{0.52\textwidth}
    \stackinset{l}{5mm}{t}{-3mm}{\huge \textbf{b}}{\includegraphics[width=\linewidth]{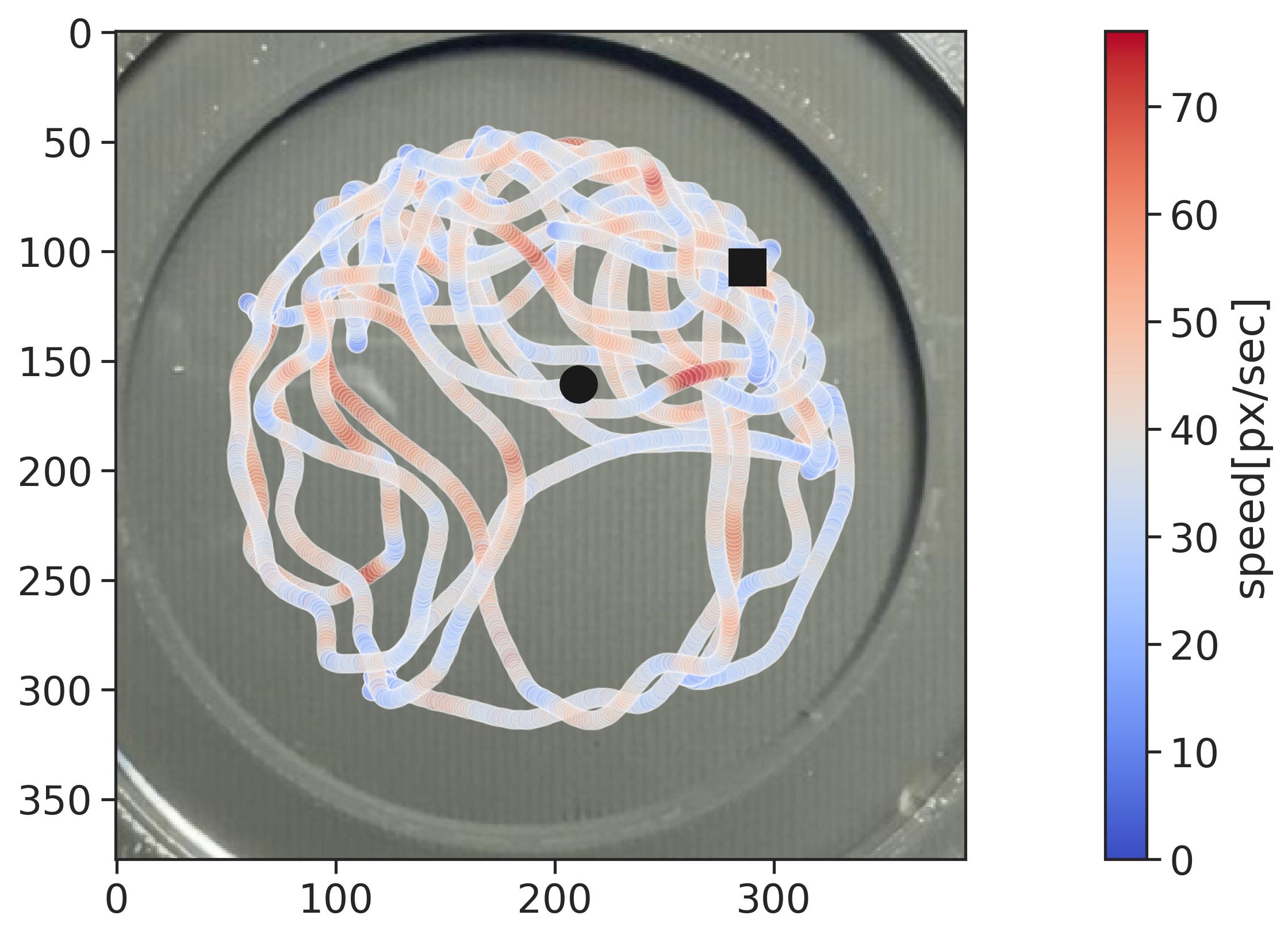}}
\end{subfigure}%

\caption{Flow \textbf{(a)} and speed map \textbf{(b)} for the Control experiment. Square and disc (black) markers represents the initial and terminal points of the motion.}
\label{fig:paths}
\end{figure}

Once the locations of the individual walkers or intruders are correctly identified, we can further analyse this data to explore certain characteristics of their motions under different experimental settings. The left figure in Fig-\ref{fig:paths} shows the location history overlaid with the flow of the motion regarding the Control experiments. We can also inspect the average speed of the droplets calculated based on its location between two consecutive detection times and create a heatmap based on this information. This is seen in the right figure in (see Fig-\ref{fig:paths}). We can observe that the droplet undergoes rapid acceleration multiple times during its motion. 

\begin{figure}[!ht]
    \centering 
\begin{subfigure}{0.461\textwidth}
  \stackinset{l}{2mm}{t}{2mm}{\huge \textbf{a}}{\includegraphics[width=\linewidth]{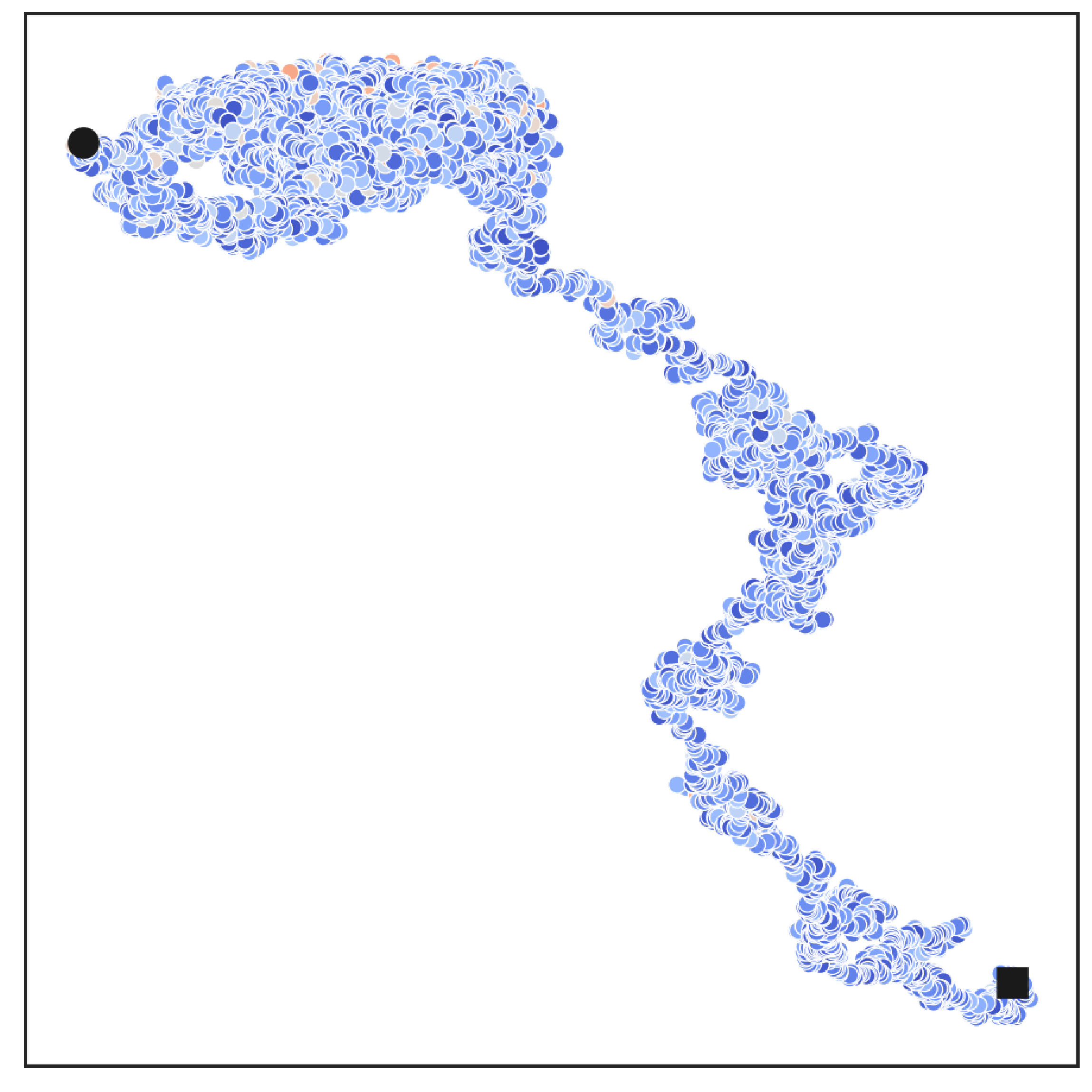}}
\end{subfigure}\hspace{1mm} 
\begin{subfigure}{0.51\textwidth}
  \stackinset{l}{2mm}{t}{1mm}{\huge \textbf{b}}{\includegraphics[width=\linewidth]{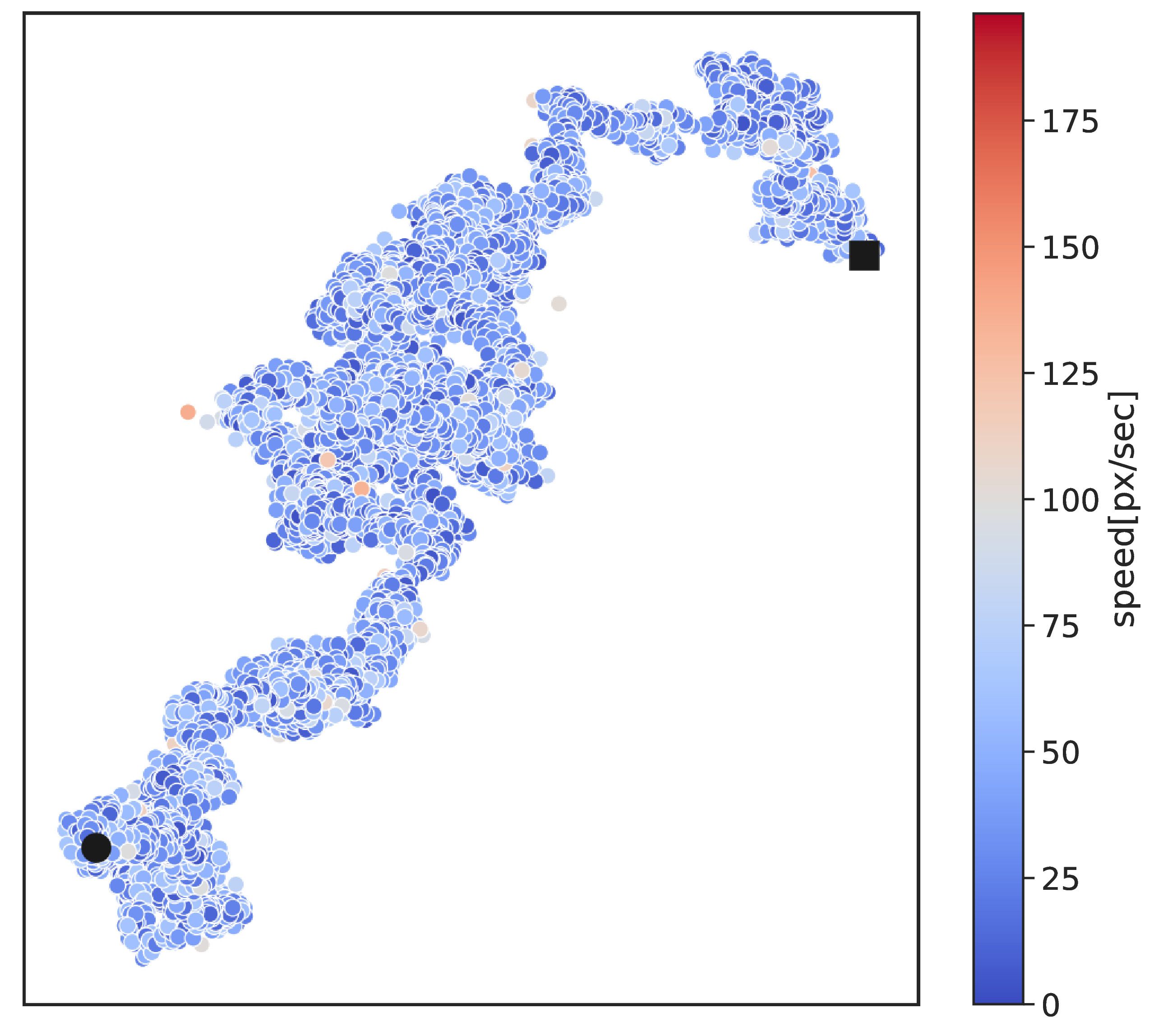}}
\end{subfigure}\hspace{1mm} 

\caption{Speed map for granular material experiment with three white intruders.  \textbf{(a, b)} represent the different intruders from 2white2black-short experiment. Square and disc (black) markers represents the initial and terminal points of the motion. Color bar representing the speed variations is shared between both figures.}
\label{fig:2w2bspeed}
\end{figure}

Similarly, we can inspect the speed map for the individual intruders. Our results can be seen in Fig-\ref{fig:2w2bspeed} related to 2white2black-short experiment. We observe that the motion of the intruders is generally free from rapid distortions.

Lastly, we would like to emphasize that once the model has been trained,  extraction of particle trajectories and subsequent analysis to determine relevant observable of interests can be achieved via live tracking. This provide a more efficient and versatile tool for investigating walking droplets and granular intruders. Live tracking enables us to extract relevant trajectory data directly from the experiments, reducing the need for extensive post-processing and storage, particularly in large-scale studies or when working with limited computational capabilities. For example, for granular intruder experiments, where very long durations (e.g., 12 hours) are of interest, it becomes increasingly impractical to save frames and process them offline. Moreover, with live tracking, one can adjust experimental parameters on-the-fly based on the observed particle behaviors. This adaptability can lead to more efficient exploration of parameter space, helping researchers optimize their experiments. Lastly, although our current study does not apply feedback control , the use of live tracking paves the way for future research that may benefit from real-time feedback. By monitoring particle trajectories live, researchers could potentially implement feedback control strategies to manipulate particle behavior or study the effects of external stimuli on the system.

\subsection{Hungarian Algorithm and 5 SOTA Models}
\label{sec:sort}

As detailed above, we utilize possibly the simplest approach to track individual particles in our problem setting. It demonstrates remarkable simplicity and interpretability, while effectively maintaining consistent ID assignment across various experimental settings in walking droplet and granular intruder experiments. This performance is crucial for our specific problem, as even a single ID switch would render the extracted particle trajectories completely unreliable. 

To substantiate our claim and to provide a fair evaluation, we thoroughly test the performance of SOTA tracking models: StrongSORT \cite{du2022strongsort}, OS-Sort \cite{cao2022observation}, Deep OC-SORT \cite{maggiolino2023deep}, BoT-SORT \cite{aharon2022bot}, and ByteTrack \cite{zhang2022bytetrack}. For further details about these models, readers are referred to the respective publications. Each of these models can operate on the detections provided by the very same YOLOv8 models we trained above. To implement these models and obtain the tracks for our experiments, we adopt their open-source Pytorch implementations in \cite{yolov8_tracking}.

For a fair evaluation, we similarly set the detection threshold regarding to 0.45. As expected, these models successfully track the droplets in all single droplet experiments. However, we observe that all of these models suffer from multiple ID switches in multiple droplet experiments and most of them similarly fail in granular intruder experiment. We provide the full summary in Fig-4 in supplementary material.
where green indicating successful tracking without any ID switch and red highlighting instances where an ID switch occurred during the tracking process.

\begin{figure*}[!b]
    \centering 
\begin{subfigure}{0.19\textwidth}
  \stackinset{l}{1mm}{t}{1mm}{\large \color{white} \textbf{a1}}{\includegraphics[width=\linewidth]{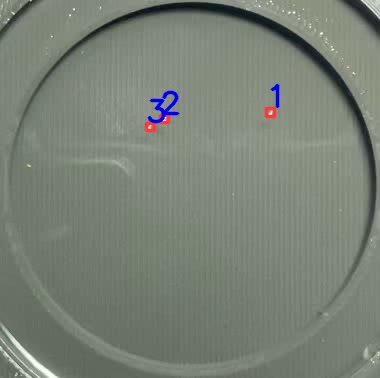}}
\end{subfigure} 
\begin{subfigure}{0.19\textwidth}
  \stackinset{l}{1mm}{t}{1mm}{\large \color{white} \textbf{b1}}{\includegraphics[width=\linewidth]{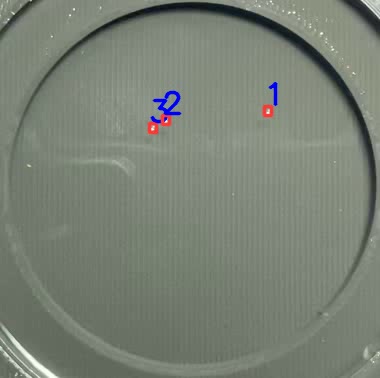}}
\end{subfigure} 
\begin{subfigure}{0.19\textwidth}
  \stackinset{l}{1mm}{t}{1mm}{\large \color{white} \textbf{c1}}{\includegraphics[width=\linewidth]{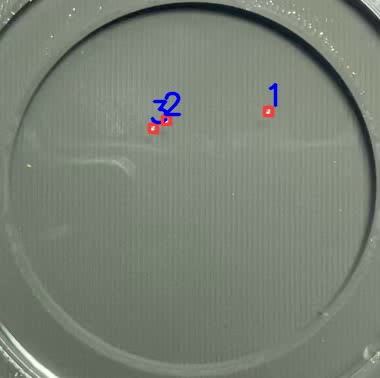}}
\end{subfigure} 
\begin{subfigure}{0.19\textwidth}
  \stackinset{l}{1mm}{t}{1mm}{\large \color{white} \textbf{d1}}{\includegraphics[width=\linewidth]{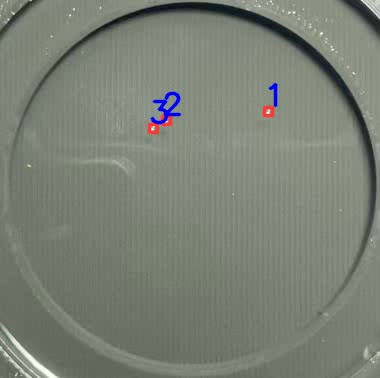}}
\end{subfigure} 
\begin{subfigure}{0.19\textwidth}
  \stackinset{l}{1mm}{t}{1mm}{\large \color{white} \textbf{e1}}{\includegraphics[width=\linewidth]{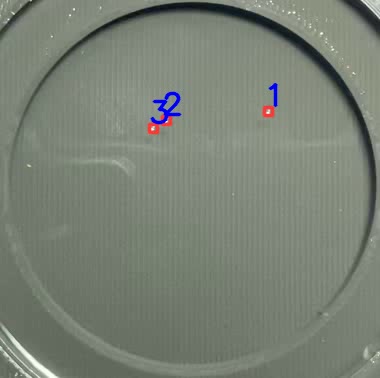}}
\end{subfigure} 

\medskip

\begin{subfigure}{0.19\textwidth}
  \stackinset{l}{1mm}{t}{1mm}{\large \color{white} \textbf{a2}}{\includegraphics[width=\linewidth]{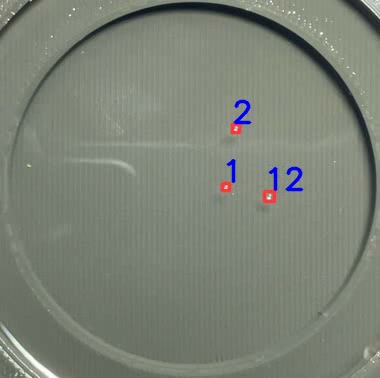}}
\end{subfigure} 
\begin{subfigure}{0.19\textwidth}
  \stackinset{l}{1mm}{t}{1mm}{\large \color{white} \textbf{b2}}{\includegraphics[width=\linewidth]{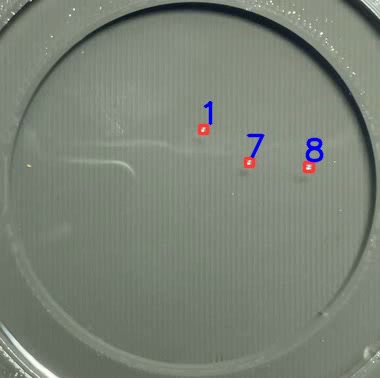}}
\end{subfigure} 
\begin{subfigure}{0.19\textwidth}
  \stackinset{l}{1mm}{t}{1mm}{\large \color{white} \textbf{c2}}{\includegraphics[width=\linewidth]{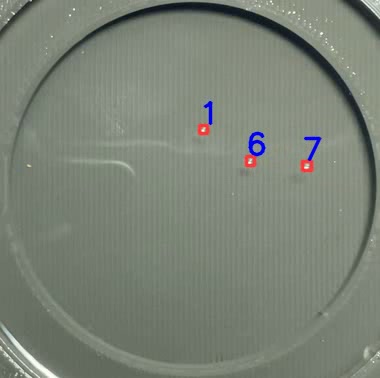}}
\end{subfigure} 
\begin{subfigure}{0.19\textwidth}
  \stackinset{l}{1mm}{t}{1mm}{\large \color{white} \textbf{d2}}{\includegraphics[width=\linewidth]{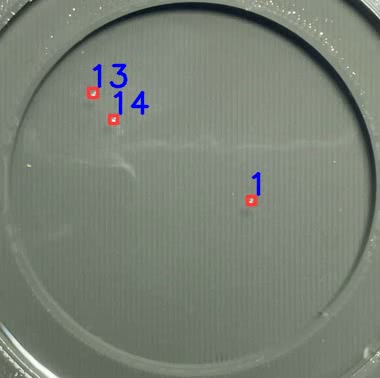}}
\end{subfigure} 
\begin{subfigure}{0.19\textwidth}
  \stackinset{l}{1mm}{t}{1mm}{\large \color{white} \textbf{e2}}{\includegraphics[width=\linewidth]{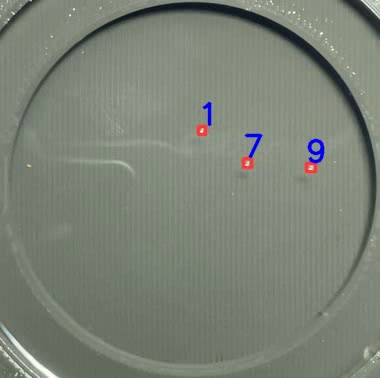}}
\end{subfigure} 
\caption{  We can see that SOTA models not only fail to maintain the initial ID assignments but also create new tracks. Initial ID assignments(top row) and ID assignments in a later frame(bottom row). StrongSORT \textbf{(a1, a2)}, OS-Sort \textbf{(b1, b2)}, Deep OC-SORT \textbf{(c1, c2)}, BoT-SORT \textbf{(d1, d2)}, and ByteTrack \textbf{(e1, e2)}.}
\label{fig:id_switch}%
 \end{figure*}

Fig-\ref{fig:id_switch} illustrates the initial ID assignments of StrongSORT, OS-Sort, Deep OC-SORT, BoT-SORT, and ByteTrack in the Three Droplet experiment. The second row of the figure depicts the corresponding ID assignments in a later frame, in the same order. We can see that SOTA models not only fail to maintain the initial ID assignments but also tend to create new tracks during the experiments.This behavior makes these models unsuitable for our experiments and also similar experiments, as even a single ID switch or the creation of a new track would render the extracted particle trajectories unreliable.


%

Results in this section demonstrate that one should be cautious prior to employing a state-of-the-art tracker to extract the trajectories in wave-particle entities experiments. The black-box nature of those methods may be particularly problematic. While our simple distance-based Hungarian Algorithm approach is free of ID switches in all experiments, most of the aforementioned SOTA model fails on numerous occasions for experiments carried out with multiple wave-particle entities despite the fact they draw upon the Hungarian algorithm. It is nontrivial to identify the source of this failure due to the black-box nature of them. Based on the high detection rates accomplished by YOLOv8, our tracking approach is interpretable in the context of our experiments. Assuming we have similar detection rates with Table-\ref{tab:ftr1}, any ID switch in our approach would most likely be caused by large displacements of wave-particle entities between two successive detections as we avoid relying on motion-predictor or feature-extractor for tracking. Recording experiments at a higher frame rate would assist our tracking algorithm in better resolving the motion of wave-particle entities. We demonstrated the viability of YOLOv8 given higher frame rates on the White Corral experiment captured at 60 fps.A similar methodology could be employed to rectify any potential ID switch that may arise within our tracking algorithm.

\section{Conclusion and Future Work}
\label{sec:conclusion}
We demonstrated a deep learning algorithm that enhances the object tracking pipeline for extracting the trajectories of objects of interest (i.e. walkers and intruders) in wave-particle entities experiments. Our tracking-by-detection pipeline uses YOLOv8 for detection and the Hungarian Algorithm for tracking. In a broad spectrum of the walking droplet and granular intruder experiments, the proposed method identifies the individual walkers and intruders with near-perfect detection accuracy and tracks them over the course of the experiment without any identity switches. Trajectory extraction thereby enables the examination of important characteristics hidden in the dynamics of the wave-particle entities.

One of the major goals of this work is to promote data-driven discovery of underlying physics governing the motion of wave-particle entities in classical experiments. An essential component of these efforts is to accurately extract the dynamics in a broad spectrum of experimental settings and to significantly automatize this process. One can then create a vast amount of rich data to serve as a testing and exploration ground for understanding these experiments. Of particular importance is understanding the resulting dynamics that emerge in $N$-particle systems driven by wave-particle dynamics. As a side note, we developed a particle simulation framework to replicate the conditions of a real walking droplet simulation by combining the trajectories we obtained in the aforementioned experiments. Our aim was to test the capacity of our model to accurately track a large number of particles in a two-dimensional space. The simulation framework was purposefully designed to capture the dynamics of particle motion, encompassing aspects such as gravitational interactions and collision avoidance. Our simulation results  demonstrate that our proposed method can effectively 10 droplets without encountering any ID-switch issues. For more detailed information, we refer readers to the Supplementary material, where this material is presented.

ur future work regarding this study is twofold. As demonstrated in Fig-\ref{fig:realtime1} and Fig-\ref{fig:realtime3}, wave-particle entities usually exhibit a complex trajectory. Therefore, we are unlikely to describe the full motion of these entities with a single set of governing equations. However, our preliminary results indicate that it could be possible to identify the governing dynamics in short patches using the sparse regression method SINDY \cite{brunton2016discovering}. This will constitute the first direction of our future work. The second direction is to investigate the existence of spatio-temporal modes which dominate the evolution of the system. This investigation is based on the hypothesis that the dynamics of this high-dimensional dataset may be described by underlying lower-dimensional patterns. To uncover these patterns, we will investigate several modal decomposition techniques such as proper orthogonal decomposition \cite{chatterjee2000introduction} and dynamic mode decomposition (DMD) \cite{kutz2016dynamic,schmid2010dynamic}.

The method presented in this manuscript will resolve numerous issues concerning tracking the long-time trajectory statistics of wave-particle entities. Even small discrepancies at inopportune times can accumulate to produce incorrect scientific results. Our present pipeline is robust across variations in experimental settings and across complex interactions among several wave-particle entities. We foresee that an improvement in the accuracy of observations will lead to better reproducibility and more transparency in experimental studies.

\section*{Acknowledgements}

The authors acknowledge support from the National Science Foundation AI Institute in Dynamic Systems (grant number 2112085).
JNK further acknowledges support from the Air Force Office of Scientific Research (FA9550-19-1-0011).

\section*{Conflict of interest}
The authors declare that they have no conflict of interest.

\vspace{-15pt}
\section*{Data Availability Statement}
The codes, datasets, and results of this study are available in GitHub repository \cite{ourgit}. We also provide detailed step-by-step tutorial on how to adopt our repository for similar problem domains. Supplementary material is also available in the same repository. 

\newpage
\bibliographystyle{unsrt}
\bibliography{dropletref}

\end{document}